\documentclass[10pt,twocolumn,letterpaper]{article}

\usepackage[pagenumbers]{cvpr} %

\definecolor{cvprblue}{rgb}{0.21,0.49,0.74}
\usepackage[pagebackref,breaklinks,colorlinks,allcolors=cvprblue]{hyperref}
\usepackage{multirow}

\newcommand{\tech}{\textsc{Rattan}}

\def\paperID{17041} %
\def\confName{CVPR}
\def\confYear{2025}

\title{Exploiting Watermark-Based Defense Mechanisms in Text-to-Image Diffusion Models for Unauthorized Data Usage}

\author{
Soumil Datta, Shih-Chieh Dai, Leo Yu, Guanhong Tao \\
University of Utah \\
Salt Lake City, UT \\
{\tt\small \{soumil.datta, shihchieh.dai, zhuoxi.yu, guanhong.tao\}@utah.edu}
}

\begin{document}
\maketitle
\begin{abstract}
Text-to-image diffusion models, such as Stable Diffusion, have shown exceptional potential in generating high-quality images. However, recent studies highlight concerns over the use of unauthorized data in training these models, which may lead to intellectual property infringement or privacy violations. A promising approach to mitigate these issues is to apply a watermark to images and subsequently check if generative models reproduce similar watermark features. In this paper, we examine the robustness of various watermark-based protection methods applied to text-to-image models. We observe that common image transformations are ineffective at removing the watermark effect. Therefore, we propose \tech{}, that leverages the diffusion process to conduct controlled image generation on the protected input, preserving the high-level features of the input while ignoring the low-level details utilized by watermarks. A small number of generated images are then used to fine-tune protected models. Our experiments on three datasets and 140 text-to-image diffusion models reveal that existing state-of-the-art protections are not robust against \tech{}.

\end{abstract}
    
\section{Introduction}
\label{sec:intro}

\begin{figure*}[t]
    \centering
    \includegraphics[width=0.85\textwidth]{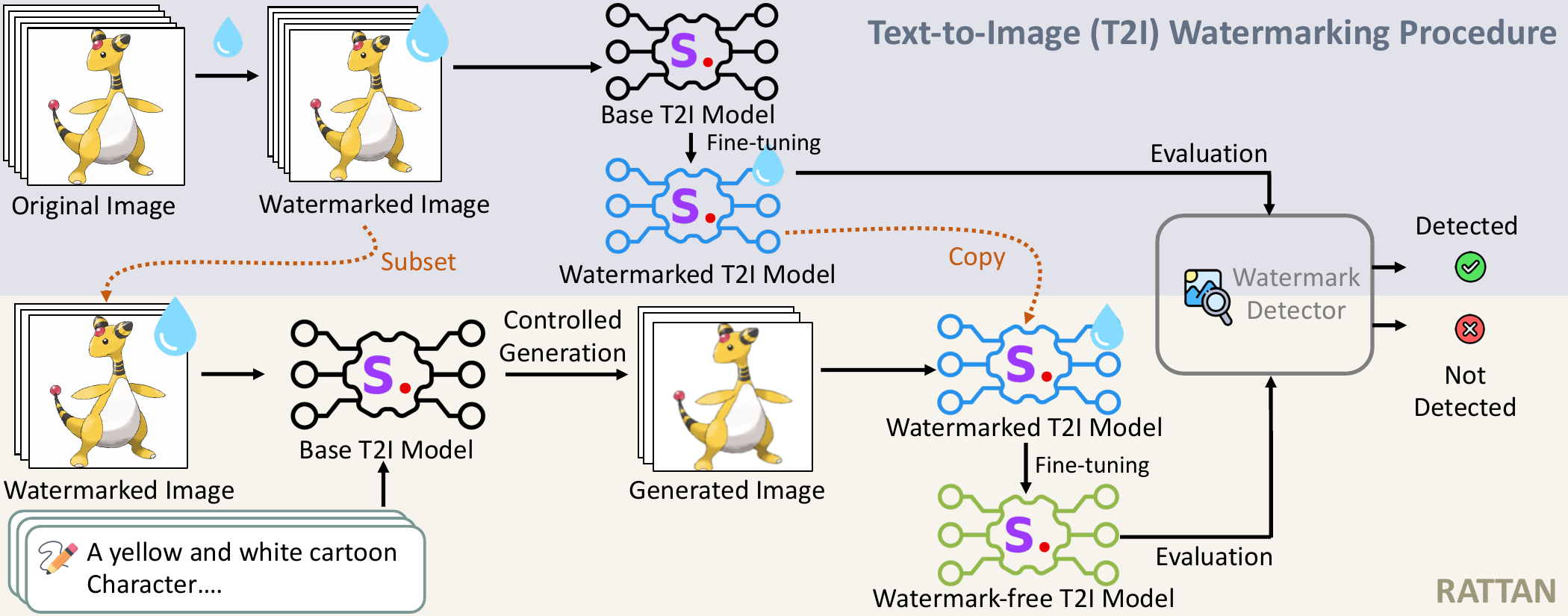}
    \caption{The top part represents the existing watermarking procedure for text-to-image diffusion models. The bottom part illustrates our method, \tech{}, for bypassing watermark-based protections. The text-to-image diffusion models shown in \textbf{black}, \textcolor{Cerulean}{\textbf{blue}}, and \textcolor{LimeGreen}{\textbf{green}}, denote the \textbf{off-the-shelf base}, \textcolor{Cerulean}{\textbf{watermarked}}, and \textcolor{LimeGreen}{\textbf{watermark-free}} versions, respectively.}
    \label{fig:pipeline}
\end{figure*}

In the rapidly evolving landscape of artificial intelligence (AI), generative AI has emerged as one of the most transformative areas~\cite{kumar_comprehensive}, with Text-to-Image (T2I) models such as Stable Diffusion~\cite{stablediffusion} and DALL-E~\cite{dalle} gaining popularity. These models have made significant strides in generating highly realistic images~\cite{yang2024pixel,Cheng_2023_WACV,nichol2021glide}, sometimes to the extent that even humans struggle to distinguish AI-generated content from actual photographs~\cite{tariang2024synthetic, bray2023testing}. Advancements in these models offer substantial benefits, including enhanced creative flexibility~\cite{wu-2022-creative,feng2024ccedit} and a reduction in manual effort~\cite{li2023diffusionmodelsimagerestoration,Rahman_2023_CVPR,dunkel2024generative}.

The success of these T2I models relies heavily on the large amounts of data available for training. For example, widely used datasets, such as the LAION dataset, contain more than 5 billion image-text pairs~\cite{schuhmann2022laion}. However, these datasets may also include images that are intellectual property or private, which might be unintentionally incorporated into the model or even intentionally exploited by an adversary~\cite{andersen2023,lu2024disguised,somepalli2023understanding,lu2024disguised}. For instance, an attacker might maliciously collect artwork from an artist without proper permission and train a T2I model on these samples. Once trained, the attacker could sell the generated images, which would be visually similar to the original artworks~\cite{cetinic2022understanding,gillotte2019copyright}. This significantly raises concerns regarding unauthorized data usage, as it leads to intellectual property infringement or privacy violations~\cite{bendel2023image,zhang2024copyright}.

To address this problem, existing research has introduced various detection and protection approaches. Membership inference attacks~\cite{shokri2017membership, carlini2023extracting,matsumoto2023membership} were originally designed to extract private information from a machine learning model by determining whether a given private input was part of its training data. These techniques can also be adapted to detect unauthorized data usage, as they share a similar goal~\cite{Dubinski_2024_WACV,li2024unveiling,pang2023black}. However, it has been shown that membership inference attacks are less effective for T2I diffusion models~\cite{duan2023diffusion}, achieving only around a 66\% detection rate on Stable Diffusion v1.5~\cite{stablediffusion}.

Another line of research leverages a watermark as a secret key to modify protected images~\cite{yu2021artificial,luo2023steal,cui2023diffusionshield,li2022untargeted}. Any T2I models trained on these samples will also ``memorize'' the watermark. During inference, the T2I model will generate images containing the watermark, which can thus be successfully detected. DIAGNOSIS~\cite{wang2023diagnosis}, a state-of-the-art watermark-based protection method, applies a stealthy coating to protected images using a specialized function. Models trained on these coated images will produce outputs with a similar coating effect, thereby flagging the use of unauthorized data.
Note that other watermarking methods aim to watermark AI-generated images~\cite{jiang2023evading,lukas2024leveraging,saberi2024robustness,zhao2024invisible}. These methods differ from the focus of this paper, as they are applied to images already generated by models and aim to identify AI-generated content. In contrast, our work focuses on preventing unauthorized data usage in T2I diffusion models, where the images in question are real (not AI-generated) and need protection from being illegitimately learned by such models.

In this paper, we investigate the robustness of state-of-the-art watermarking frameworks for T2I diffusion models. Specifically, we apply a set of common image transformations to watermarked images and observe that these transformations are largely ineffective at removing the watermark effect.
This is due to the design of these approaches, which aim to preserve the fine-grained details of inputs -- precisely the space utilized by watermarks. However, the primary goal of training on protected data is to generate images with similar high-level features. Hence, we propose to extract high-level coarse-grained features from protected images while ignoring low-level details. Particularly, we introduce \tech{} (\underline{\textbf{R}}emoving w\underline{\textbf{AT}}ermarks in \underline{\textbf{T}}ext-to-im\underline{\textbf{A}}ge diffusio\underline{\textbf{N}} models), which applies a controlled image generation process to protected images. The bottom part of \autoref{fig:pipeline} illustrates the pipeline of \tech{}. Specifically, we leverage an off-the-shelf T2I model to generate a new version of the input based on both the protected image and the corresponding text. \tech{} takes advantage of the diffusion process to preserve key input features while removing unnecessary low-level details. The generated images are then used to fine-tune the model. Our evaluation on three datasets and 140 text-to-image diffusion models demonstrates that \tech{} can significantly reduce the detection rate of existing protections to 50\%, equivalent to random guessing.

\section{Background and Related Work}
\label{sec:background}

\subsection{Text-to-Image Diffusion Model}
Text-to-image diffusion models have gained popularity for their ability to generate high-quality images from textual descriptions~\cite{dhariwal2021diffusion, saharia2022photorealistic, zhang2023adding}. Stable Diffusion~\cite{stablediffusion}, like other models, is an open-source generative model that iteratively transforms noisy data into coherent images through a diffusion process, producing highly detailed and diverse outputs~\cite{10156981, kingma2021variational}.

The diffusion process can be intuitively understood as a procedure of sequentially adding and removing Gaussian noise. Starting with an input image, random noise sampled from a normal distribution is iteratively added until the image is transformed into pure Gaussian noise. The goal of training a diffusion model is to learn a neural network that reverses this process, iteratively removing noise until the original input is recovered \cite{yang2023diffusion, croitoru2023diffusion, cao2024survey}. Formally, in the forward process of adding noise, given an initial input $x^0$, its value at time step $t$ is:
\begin{equation}
    x^t = \alpha^t \cdot x^0 + \sigma^t \cdot z, z \sim \mathcal{N}(\mathbf{0}, \mathbf{I}),
    \label{eq:diffusion_add_noise}
\end{equation}
where $\alpha^t$ is a constant derived from $t$ that denotes the magnitude of $x^t$, and $\sigma^t$ is a constant derived from $t$ that determines the magnitude of noise $z$ added to the image. The noise $z$ is sampled from a standard normal distribution.
The training objective of diffusion models is, therefore, to minimize the difference between the initial input $x^0$ and the denoised output obtained from $x^t$ after processing it through the model multiple times.
\begin{align}
    &\min_\theta ||\hat{x}^{t-1} - x^0||_2^2,\\
    &\hat{x}^{t-1} = \hat{x}^t + \epsilon^2 \cdot f_\theta (\hat{x}^t, t) + \epsilon \cdot z, \label{eq:diffusion_denoise}
\end{align}
where $\hat{x}^t$ starts from $x^t$. Here, $\epsilon$ is a constant derived from $\sigma$, and $f_\theta$ represents the diffusion model, which takes the input $\hat{x}^t$ and the current denoising time step $t$.

Stable Diffusion incorporates information extracted from text into the denoising process. Specifically, $f_\theta$ takes an additional feature vector representing the text features, allowing it to generate images that align with the descriptions provided in the input text.

However, training these models from scratch requires substantial computational resources. Consequently, most research focuses on fine-tuning pre-trained models~\cite{moon2022fine,shen2023finetuning, fan2024reinforcement}, enabling them to adapt to specific tasks or datasets with a fraction of the computational power otherwise required.

\subsection{Detecting Unauthorized Data Usage}

Numerous research efforts have focused on detecting the use of unauthorized data in training and on mitigating models' tendencies to memorize sensitive or copyrighted information. Membership inference attacks~\cite{hayes2017logan, duan2023diffusion} are a feasible method to potentially determine if models have memorized data from the training set. Other defense mechanisms ~\cite{shan2023glaze, van2023anti, liu2024metacloak} introduced perturbations to images so that the model would be unable to learn specific features or styles from the image, thus preventing it from reproducing the image later. Similarly, backdoor watermarking techniques perturb the input training set with a watermark, such as injected steganography-based approaches that can embed binary strings into the image~\cite{yu2021artificial, luo2023steal}, or backdoor image-transformation-based methods\cite{wang2023diagnosis}, so the model can learn and produce the watermark in its generated images. DIAGNOSIS~\cite{wang2023diagnosis} is a state-of-the-art watermark-based protection against unauthorized data usage in text-to-image diffusion models. It considers two watermark scenarios: the unconditional watermark, where the watermark is always present regardless of the text prompt, and the trigger-conditioned watermark, which appears only when a specific trigger sequence is included in the prompt.

\section{Threat Model}
\label{sec:threatmodel}

In this work, we focus on evaluating the performance of watermark-based protections against unauthorized data usage in text-to-image diffusion models. This scenario involves two parties: the data owner or a third-party data protector, who acts as the defender, and the unauthorized model developer, who is the adversary.

\smallskip
\noindent
\textbf{Data Owner or Third-Party Data Protector (Defender).}
The defender's goal is to prevent unauthorized usage or abuse of specific images that are either intellectual property or private. To achieve this, the data owner or a third-party data protector can intentionally embed imperceptible watermarks in these images, which serve as a secret key known only to the defender. Once text-to-image diffusion models are trained on such watermarked data, they will generate images containing the watermark. Thus, the defender's goal is to \textit{detect whether a given model has been trained on unauthorized data} by inspecting its generated images. The data owner or protector has full access to the protected data but not to other training data used by the diffusion model. They do not have access to the model itself or the training process and can only query the trained model to obtain generated images. This setup simulates scenarios involving model API providers, such as Midjourney~\cite{midjourney}.

\smallskip
\noindent
\textbf{Unauthorized Model Developer (Adversary).}
The adversary, or unauthorized model developer, aims to use protected images to train their text-to-image diffusion model so that it can generate similar images or images with specific features. However, the adversary wants to avoid the detection of unauthorized data usage in their trained model. To achieve this, they may first clean the training data, such as applying image transformations, before using it to train the text-to-image model. They might also inspect or modify the model post-training to remove any embedded watermarks. The adversary does not know whether or which images contain watermarks, nor do they have access to the watermark detector developed by the defender to determine if the model has been trained on unauthorized data.

\section{Method}
\label{sec:method}

The main characteristic of watermark-based protections is that the added coatings or perturbations are visually imperceptible. This ensures that the watermark, or ``secret key,'' is known only to the data owner and remains hidden from data consumers, such as text-to-image model developers. Given this, our goal is to evaluate the robustness of these watermarks -- for example, to determine if they can be removed or degraded by various image transformations.

\subsection{Image Transformation}
\label{subsec:transformation}

\begin{figure*} 
    \centering
    \includegraphics[width=0.85\textwidth]{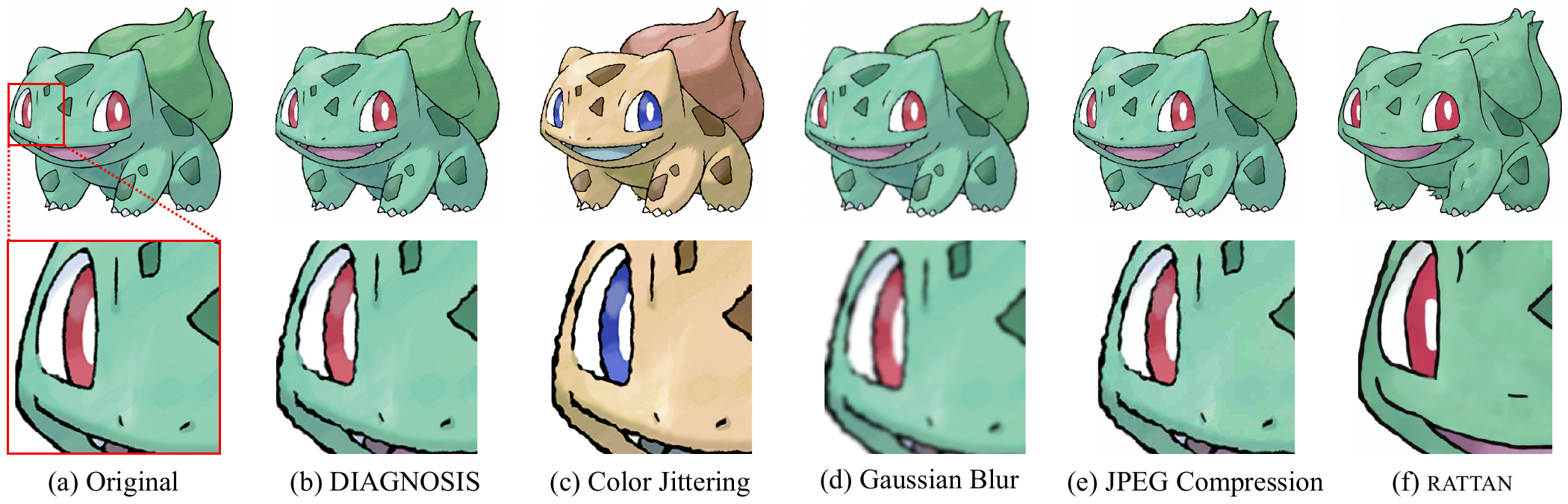}
    \caption{Comparison of (a) the original image, (b) DIAGNOSIS watermarked image, and the images after applying (c) Color Jittering, (d) Gaussian Blur, (e) JPEG Compression. The bottom row provides a zoomed-in view. The curly-line characteristic of the watermark is still visible in each transformed image. (f) presents the result after \tech{}'s controlled image generation on the watermarked input. The lines appear smoother compared to the original image.} \label{fig:transform-images}
\end{figure*}

Since watermark-based protections add small perturbations to protected images, a straightforward approach to evaluate their robustness is to apply various image transformations to these images. We leverage three commonly used transformation methods: Gaussian blur, JPEG compression, and color jittering.
\autoref{fig:transform-images} shows the images after applying the aforementioned transformation methods. The second column presents the image embedded with the DIAGNOSIS watermark, which is an image-warping function that distorts straight lines into curly ones, as illustrated in image (b) compared to the original image (a). After applying the different transformations, it can be observed that the boundary lines remain curly, indicating that the watermark effect persists.
We further evaluate six other image transformations in \autoref{subsec:simple-transform} and reach the same conclusion.

Image transformations fail to remove the watermark effect because they cannot significantly alter the input image (to preserve the primary content). These watermarks are designed to be robust against common distortions. For instance, an attacker might photograph a piece of art on display and use that image for model training. The watermark should persist despite adjustments in lighting, contrast, noise, and other minor variations. Consequently, simple image transformations cannot effectively eliminate the watermark effect. However, this does not imply that existing protections are immune to all possible manipulations. In the following section, we demonstrate how our new design can bypass these protections.

\subsection{Our Solution}
\label{subsec:our_solution}

The primary function of text-to-image models is to generate images that align with input text descriptions. When unauthorized data is used in training, the model’s capability extends to generating images with features similar to those in the protected images. Existing watermark-based protections assume that for a model to generate similar images, it must be trained on the exact content of the protected images. This process would lead to the watermark being embedded into the model. However, this assumption does not always hold. As long as the model learns the key features of the protected data, it can generate similar images without directly replicating the original content (e.g., the watermark).

We propose to leverage the generative capabilities of diffusion models to construct data samples that share key features with protected images. Instead of directly using the protected images in training, we employ a technique similar to zero-shot learning. Here, a diffusion model is prompted to generate an image based on a text description along with a reference (protected) image. In this way, the diffusion model has the freedom to create images without focusing on details such as the watermark embedded in the protected images.

\autoref{fig:pipeline} presents an overview of our method, \tech{}, for bypassing watermark-based protections. The top part of the figure illustrates how existing watermark-based protection methods operate, which involve two steps. The first step is embedding a watermark onto protected images. This watermark can be either a sequence of pixel value bits or an image transformation function. The second step involves inspecting the generated images from text-to-image diffusion models. If the model has been trained on watermarked data, the generated images will also contain this watermark. Consequently, existing protections can flag the trained model.

The bottom part of \autoref{fig:pipeline} shows the pipeline of our technique, \tech{}. The adversary’s goal is to train a text-to-image model on unauthorized data. Without knowing whether these protected images are watermarked, \tech{} selects a subset of the data (e.g., 10 images) and their corresponding text descriptions. It then employs an off-the-shelf Stable Diffusion~\cite{stablediffusion} model to perform controlled image generation based on both the input image and its corresponding text. The generated images are intended to retain the high-level key features of the protected data but be free of watermarks. \tech{} then uses this small set of images to fine-tune the potentially watermarked text-to-image model.
Below, we detail the two main components of \tech{}: controlled image generation and watermark removal.

\smallskip
\noindent
\textbf{Controlled Image Generation.}
As discussed above, our goal is to obtain images that preserve key features of protected data without copying the fine-grained details used by watermarks. Our idea is to leverage an off-the-shelf diffusion model to perform controlled image generation. Specifically, the model has a certain freedom to create an image based on the given text and the protected input.

For diffusion models, the generation process typically starts from random Gaussian noise, as described in \autoref{eq:diffusion_denoise}, and then iteratively removes the noise by passing it through the model. In our scenario, we aim to generate an image that retains the major features of the protected input, such as structures and outlines. To achieve this, similar to existing work~\cite{mengsdedit}, rather than starting the generation process from random Gaussian noise, \tech{} uses a starting point obtained from the protected image.

Specifically, given the projected image $x_{protected}$, we first apply \autoref{eq:diffusion_add_noise} to it, which essentially adds random noise to the image.
\begin{align}
    x^t_{guide} &= \alpha^t \cdot x^{t-1}_{guide} + \sigma^t \cdot z, z \sim \mathcal{N}(\mathbf{0}, \mathbf{I}), \label{eq:diffusion_pretected}\\
    x^0_{guide} &= x_{protected}
\end{align}
In the standard diffusion process, the output from the above equations after $t$ iterations becomes Gaussian noise, and the diffusion model aims to recover the original input. This is why, after training, the diffusion model can generate images from random noise. Here, our goal is to preserve the key high-level features of the protected input, such as structures, outlines, color schemes, etc., so that the diffusion model can recover these coarse features while filling in fine-grained details. Thus, we do not add noise to $x_{protected}$ until it becomes Gaussian noise but rather stop at a certain step. Suppose the total number of iterations needed to transform an image to random noise is $t$; our diffusion process (\autoref{eq:diffusion_pretected}) only applies for $\gamma \cdot t$ iterations. Empirically, we choose $\gamma = 0.6$ as it provides the best trade-off between the quality of generated images and the evasion rate. The results of using different $\gamma$ values are discussed in \autoref{sec:ablation}.

After obtaining the diffused $x_{protected}$, i.e., $x_{guide}$, we pass it through a standard diffusion model to generate a new image, as illustrated in \autoref{eq:diffusion_denoise}. Note that we use an off-the-shelf diffusion model (not the model trained on the protected data) in \tech{}, with its weight parameters frozen.

In addition to the original protected image, we also include its paired text description as a reference. This is because the final text-to-image model is trained on both the image and the text, with the text providing guidance on which parts the model should focus on during training. Thus, we leverage the text in our controlled image generation as well. This approach follows the standard Stable Diffusion~\cite{stablediffusion} inference process, where the text embedding from a text encoder is incorporated into the cross-attention layers during the denoising process. More details can be found in the original paper~\cite{stablediffusion}.

The last column (f) in \autoref{fig:transform-images} shows the result after applying \tech{} to the watermarked input (b). Observe that the generated image has smooth boundary lines, effectively removing the watermark effect present in (b). Additionally, the Pokémon's teeth are no longer visible, and the color tone differs from the original image. This is due to the controlled generation process, which preserves high-level features while disregarding low-level details.

\smallskip
\noindent
\textbf{Watermark Removal.}
Since our generated images from protected inputs do not contain watermarks, a straightforward approach is to apply controlled image generation to all the training data. However, there are two issues with this method. First, the training set could be extensively large, and applying the diffusion model generation to all samples can be computationally expensive and substantially increase costs. Second, as discussed earlier, the controlled generation retains the key high-level features of protected images while ignoring low-level details. Although this helps to eliminate watermarks, it also removes fine-grained features necessary for training text-to-image models.

The watermarked model has already been trained on protected images with fine-grained details, including the watermark. We only need to remove the watermark without affecting the fine-grained content features. To achieve this, we propose to fine-tune the watermarked model on a small set of our generated images. Note that the text-to-image model is trained on text-image pairs. The original watermarked model has learned the correspondence between the text and the watermarked image. We use the same text but pair it with our generated image -- a slightly different version of the protected image. This guides the model to ignore the watermark and focus on the main content features, both coarse-grained and fine-grained. Our evaluation in \autoref{subsec:evading-protections} shows that with as few as 10 images, \tech{} can effectively eliminate the watermark effect.

\section{Evaluation}
\label{sec:evaluation}
This section discusses the evaluation on \tech{} and a few image transformations, as well as ablations studies to understand the impact of different components in \tech{}.

\subsection{Experimental Setup}
\label{subsec:experimental-setup}

\noindent
\textbf{Datasets.}
We utilize three popular datasets: Pokemon~\cite{pokemondataset} (833 text-image pairs), Naruto~\cite{narutodataset} (1221 text-image pairs), and CelebA~\cite{liu2015deep}. For the CelebA dataset, we use the first 1000 text-image pairs to be consistent with the experimental setting in the DIAGNOSIS paper.

\smallskip\noindent
\textbf{Watermarking Methods.}
We use three watermark-based protection methods for evaluation, namely the works by Luo \etal~\cite{luo2023steal}, Yu \etal~\cite{yu2021artificial}, and DIAGNOSIS~\cite{wang2023diagnosis}. Luo \etal and Yu \etal employ a bit string as the watermark and DIAGNOSIS uses a warping function.
DIAGNOSIS supports both unconditional and trigger-conditioned watermarks. In the unconditional setting, the watermark is activated for any text prompt. In the trigger-conditioned setting, the watermark appears only when the prompt contains a specific text trigger or a predefined token sequence at the beginning.

\smallskip\noindent
\textbf{Models and Fine-tuning.}
We primarily use Stable Diffusion v1.4~\cite{stablediffusion} along with the Low-Rank Adaptation of Large Langauge Models (LoRA) fine-tuning method~\cite{hu2021lora} for our experiments. The evaluation on other diffusion models is presented in the supplementary. We also include ablation studies with different Stable Diffusion models for controlled generation in the \tech{} pipeline in \autoref{sec:ablation}. For each experiment, we train 10 models for both clean and watermarked models to minimize the randomness.

\smallskip\noindent
\textbf{Metrics.}
We make use of several metrics to evaluate the results. For the detection on unauthorized data usage, we use True Positives (malicious models detected as malicious), True Negatives (benign models detected as benign), False Positives (benign models detected as malicious), and False Negatives (malicious models detected as benign).
For example, the goal of \tech{} is to shift the results of True Positives towards False Negatives, i.e., allow malicious models with unauthorized data usage to be considered benign.
We also calculate the average of Fréchet Inception Distance (FID)~\cite{heusel2018ganstrainedtimescaleupdate} of the generated images from each model to measure the generation quality, as well as report on the model's memorization strength to determine how close it is to being detected as malicious.

\begin{table}[t]
    \centering
    \footnotesize
    \caption{Detection results of different watermark-based protections before and after applying \tech{}.}
    \begin{tabular}{llccccc}
        \toprule
        Watermark & Method & TP & TN & FP & FN & Acc. \\
        \midrule
        \multirow{2}{*}{Luo \etal~\cite{luo2023steal}} & Original & 0 & 5 & 0 & 5 & 50\% \\
                    & \tech{} & 0 & 5 & 0 & 5 & 50\% \\
        \midrule
        \multirow{2}{*}{Yu \etal~\cite{yu2021artificial}} & Original & 0 & 5 & 0 & 5 & 50\% \\
                   & \tech{} & 0 & 5 & 0 & 5 & 50\%\\
        \midrule
        \multirow{2}{*}{DIAGNOSIS~\cite{wang2023diagnosis}} & Original & 5 & 5 & 0 & 0 & 100\% \\
                   & \tech{} & 0 & 5 & 0 & 5 & 50\% \\
        \bottomrule
    \end{tabular}
    \label{tab:different_watermarks}
\end{table}

\subsection{Evading Watermark-based Protections}
\label{subsec:evading-protections}

In this section, we evaluate the performance of \tech{} on the works by Luo \etal~\cite{luo2023steal}, Yu \etal~\cite{yu2021artificial}, and DIAGNOSIS~\cite{wang2023diagnosis}. \autoref{tab:different_watermarks} shows the results on the Pokemon dataset. For each watermarking method, we trained 10 models: 5 models with the watermark in the training set and 5 benign models where the training set is unmodified. As shown in the table, the watermarks implemented by Luo \etal and Yu \etal are not effectively memorized by the diffusion model, making them unreliable for detecting malicious behavior.
Luo \etal and Yu \etal’s methods implement fingerprinting of images using binary strings, with detection conducted by measuring the bit accuracy of generated samples. The average bit accuracy for raw images (images with the watermark directly added) produced by Luo \etal was approximately 68.11\%, while for Yu \etal, it was around 46.93\%. This demonstrates the limited effectiveness of these methods in embedding watermarks.
In contrast, DIAGNOSIS performs significantly better, with a 100\% detection accuracy. However, for all three watermarking methods, \tech{} successfully reduces their detection accuracy to 50\%, equivalent to random guessing.

\smallskip\noindent
\textbf{Results on DIAGNOSIS.}
As observed above, since the other two watermarking methods result in poor memorization of watermarks in diffusion models, we mainly focus on DIAGNOSIS in the following evaluation.

For our experiments, we adopt a similar setting to the one used in the DIAGNOSIS paper. We use 50 different text prompts to generate images from the fine-tuned models and report the FID scores and memorization strengths. We train 20 models for each case: 10 models using unauthorized data and 10 models without unauthorized data. Our goal is to shift True Positives towards False Negatives, achieving an overall accuracy of 50\%, equivalent to random guessing. \autoref{tab:evaluation-methods} presents the evaluation results.

\begin{table*}[t]
    \centering
    \footnotesize
    \caption{Evaluation of \tech{} against DIAGNOSIS across various datasets.}
    \begin{tabular}{cccccccccc}
        \toprule
        \multirow{2}{*}{\textbf{Dataset}} & \multirow{2}{*}{\textbf{Method}} & \multirow{2}{*}{\textbf{Memorization Type}} & \multicolumn{5}{c}{\textbf{Metrics}} & \multirow{2}{*}{\textbf{FID $\downarrow$}} & \multirow{2}{*}{\textbf{Memorization $\downarrow$}} \\
        \cmidrule(lr){4-8}
        & & & TP & TN & FP & FN & Acc &  & \\
        \midrule
        \multirow{4}{*}{Pokemon}
            & \multirow{2}{*}{Diagnosis} 
            & \renewcommand{\arraystretch}{1.5}\hspace{1mm}Unconditional\renewcommand{\arraystretch}{1} & 10 & 10 & 0 & 0 & 100\% & 214.86 $\pm$ 9.02 & 0.830 \\
            & & Trigger-Conditioned & 10 & 10 & 0 & 0 & 100\% & 270.57 $\pm$ 11.61 & 1.000 \\ \cmidrule(lr){2-8} \cmidrule(lr){9-10}
            \rowcolor{gray!25} \cellcolor{white}& 
            & \renewcommand{\arraystretch}{1.5}\hspace{1mm}Unconditional\renewcommand{\arraystretch}{1} & 0 & 10 & 0 & 10 & 50\% & 211.59 $\pm$ 3.15 & 0.327\\
            \rowcolor{gray!25} \cellcolor{white}& \multirow{-2}{*}{\tech} & Trigger-Conditioned & 0 & 10 & 0 & 10 & 50\% & 214.09 $\pm$ 3.33 & 0.173\\
        \cmidrule(lr){1-10}
        \multirow{4}{*}{Naruto}
            & \multirow{2}{*}{Diagnosis} 
            & \renewcommand{\arraystretch}{1.5}\hspace{1mm}Unconditional\renewcommand{\arraystretch}{1} & 7 & 10 & 0 & 3 & 85\% & 240.13 $\pm$ 7.15 & 0.790\\
            & & Trigger-Conditioned & 10 & 10 & 0 & 0 & 100\% & 257.67 $\pm$ 9.61 & 0.912\\
        \cmidrule(lr){2-8} \cmidrule(lr){9-10} %
            \rowcolor{gray!25} \cellcolor{white}& 
            & \renewcommand{\arraystretch}{1.5}\hspace{1mm}Unconditional\renewcommand{\arraystretch}{1} & 0 & 10 & 0 & 10 & 50\% & 241.91 $\pm$ 6.77 & 0.360\\
            \rowcolor{gray!25} \cellcolor{white}& \multirow{-2}{*}{\tech} & Trigger-Conditioned & 0 & 10 & 0 & 10 & 50\% & 244.82 $\pm$ 10.98 & 0.246\\
        \cmidrule(lr){1-10}
        \multirow{4}{*}{CelebA}
            & \multirow{2}{*}{Diagnosis} 
            & \renewcommand{\arraystretch}{1.5}\hspace{1mm}Unconditional\renewcommand{\arraystretch}{1} & 10 & 10 & 0 & 0 & 100\% & 237.63 $\pm$ 6.33 & 0.996\\
            & & Trigger-Conditioned & 10 & 10 & 0 & 0 & 100\% & 240.11 $\pm$ 8.99 & 1.000\\
        \cmidrule(lr){2-8} \cmidrule(lr){9-10} %
            \rowcolor{gray!25} \cellcolor{white} & 
            & \renewcommand{\arraystretch}{1.5}\hspace{1mm}Unconditional\renewcommand{\arraystretch}{1} & 0 & 10 & 0 & 10 & 50\% & 230.52 $\pm$ 5.55 & 0.491\\
            \rowcolor{gray!25} \cellcolor{white}& \multirow{-2}{*}{\tech} & Trigger-Conditioned & 0 & 10 & 0 & 10 & 50\% & 232.66 $\pm$ 4.01 & 0.299\\
        \bottomrule
    \end{tabular}
    \label{tab:evaluation-methods}
\end{table*}

Observe that DIAGNOSIS performs well on most datasets, with high accuracy and memorization strengths. However, \tech{} can successfully shift the true positives towards false negatives, yielding a 50\% detection accuracy by DIAGNOSIS, which is equivalent to random guessing. The memorization is significantly reduced from nearly 1 to 0.3 in most cases. Additionally, the FID scores of the models after applying \tech{} are similar or even lower than those of DIAGNOSIS-watermarked models. This demonstrates that \tech{} can effectively preserve the normal functionality of T2I models while removing watermarks.

\subsection{Performance of Image Transformations}
\label{subsec:simple-transform}

As discussed in \autoref{subsec:transformation}, one straightforward idea to remove watermarks is to apply image transformations. We have shown earlier that Gaussian blur, JPEG compression, and color jittering cannot remove the watermark embedded by DIAGNOSIS. Here, we evaluate six more image transformations, including saturation increase, using 8-bit quantization, adding a green hue, increasing the contrast, cropping by a factor of 1.5 on each side, and increasing the brightness. The results are shown in \autoref{tab:ablation-results}.

From the table, we observe that DIAGNOSIS is highly effective against most image transformations. However, its performance declines when contrast is increased in the training set. The strong performance of DIAGNOSIS against popular image transformation-based mitigation strategies underscores the need for \tech{} to provide a more robust evaluation.

\subsection{Visualization of \tech{} Generated Images}

\autoref{fig:transform-images} illustrates the image generated by \tech{} based on the protected input. The bottom row provides a zoomed-in view of a specific region of the image. The first column shows the original image before undergoing DIAGNOSIS's watermarking process. In the second column, DIAGNOSIS applies a watermark, noticeable as slight wobbliness along the edges. This edge perturbation introduces a feature that can be subtly learned by the diffusion model while remaining robust against most image transformations. The last column presents the result after \tech{} generates a new image based on the watermarked input and the corresponding text. \tech{}’s controlled image generation smooths out these added perturbations, effectively ignoring low-level details. Importantly, the generated image retains most of the essential features of the original image, ensuring the intellectual property quality remains unaffected.

\subsection{Ablation Study}
\label{sec:ablation}
In this section, we present the results of our ablation study on the performance of \tech{}. First, we investigate how fine-tuning with different numbers of cleaned samples affects the efficacy of our method. Next, we examine the impact of varying the number of fine-tuning epochs. Finally, we evaluate the effectiveness of our approach across various diffusion models. Additionally, we consider training a model from scratch using \tech{}-generated images. The results of each ablation study on the Pokemon dataset are shown in \autoref{tab:ablation-results}. Since the protection method does not misclassify benign models (those not trained on unauthorized data) as malicious, we only report the detection rate for watermarked models after applying \tech{} in \autoref{tab:ablation-results}.

\smallskip\noindent
\textbf{Impact of \tech-Generated Samples.}
We first examine the effect of fine-tuning on different numbers of \tech-generated images, ranging from as few as 5 samples to the entire training set of 783 samples.
The results in \autoref{tab:ablation-results} indicate that fine-tuning with a smaller subset of cleaned samples results in relatively lower FID scores.
This occurs because \tech-generated images primarily preserve high-level features while disregarding low-level details. Fine-tuning on a larger amount of such data may impact the model’s ability to generate high-quality images.
On the other hand, using the full dataset results in a few malicious detections, suggesting that training on a larger number of samples might increase the risk of overfitting to the watermarked features.

\begin{table}[t]
    \centering
    \caption{Ablation study across several factors: the image transform method, the number of fine-tuning samples, the number of epochs, parameter $\gamma$, and the version of Stable Diffusion.}
    \resizebox{\columnwidth}{!}{
    \begin{tabular}{clcccc}
        \toprule
         Ablation & & FID $\downarrow$ & Detection $\downarrow$ & Memorization $\downarrow$ \\
        \midrule \midrule
       \multirow{6}{*}{\shortstack[c]{Image \\ Transform}} & Saturation & 229.80 $\pm$ 6.15 & 90\% & 0.874 \\
        &  8-bit Quant. & 223.30 $\pm$ 7.76 & 90\% & 0.860 \\
        &  Hue Shift (Green) & 243.14 $\pm$ 8.64 & 100\% & 0.896 \\
        &  Contrast & 234.51 $\pm$ 7.83 & 70\% & 0.718 \\
        &  Cropped & 234.68 $\pm$ 7.01 & 100\% & 0.852 \\
        &  Brightness & 232.24 $\pm$ 7.39 & 90\% & 0.840 \\

        \midrule
         \multirow{6}{*}{\# Sample}&  5  & 217.97 $\pm$ 6.81 & 0\% & 0.199 \\
         &  10 & 214.27 $\pm$ 4.19 & 0\% & 0.253\\
         &  50 & 209.36 $\pm$ 4.34 & 0\% & 0.120 \\
         &  200 & 210.76 $\pm$ 6.75 & 0\% & 0.132 \\
         &  500 & 215.29 $\pm$ 7.38 & 0\% & 0.141 \\
         &  783 & 234.62 $\pm$ 3.91 & 20\% & 0.640 \\

        \midrule
         \multirow{5}{*}{\# Epoch}&  5 & 214.66 $\pm$ 4.80 & 0\% & 0.336 \\
         &  15 & 220.10 $\pm$ 7.26 & 0\% & 0.296 \\
         &  30 & 220.53 $\pm$ 5.82 & 0\% & 0.284 \\
         &  50 & 211.99 $\pm$ 5.82 & 0\% & 0.224 \\
         &  100 & 214.24 $\pm$ 3.24 & 0\% & 0.168 \\

        \midrule
         \multirow{5}{*}{$\gamma$}&  0.2 & 234.15 $\pm$ 11.77 & 80\% & 0.830 \\
         &  0.4 & 226.29 $\pm$ 6.59 & 20\% & 0.560 \\
         &  0.6 & 211.59 $\pm$ 3.15 & 0\% & 0.327 \\
         &  0.8 & 218.95 $\pm$ 6.99 & 0\% & 0.518 \\
         &  1.0 & 227.72 $\pm$ 5.56 & 50\% & 0.744 \\

        \midrule
         \multirow{3}{*}{Model}&  SD v1.4  & 210.70 $\pm$ 8.65 & 0\% & 0.193 \\
         &  SD v2.0 & 216.24 $\pm$ 6.40 & 0\% & 0.224 \\
         &  SD v3 Medium & 216.46 $\pm$ 6.38 & 0\% & 0.202 \\
        \bottomrule
    \end{tabular}}
    \label{tab:ablation-results}
\end{table}

\smallskip\noindent
\textbf{Impact of Fine-tuning Epochs.}
In this experiment, we examine the impact of varying the fine-tuning epochs in the \tech{} pipeline. By testing a range of epochs from 5 to 100, we evaluate the trade-off between image quality, as measured by the FID score, and the degree of memorization.
The results in \autoref{tab:ablation-results} indicate that the number of fine-tuning epochs has a relatively minor impact on the FID score. However, memorization strength decreases significantly as the number of epochs increases, dropping from 0.336 at 5 epochs to 0.168 at 100 epochs. This decline suggests that extended fine-tuning progressively reduces the retention of watermarked content, resulting in more effective watermark removal. However, it is worth noting that longer fine-tuning periods are computationally expensive. To achieve a balance between watermark removal efficacy and computational efficiency, \tech{} employs 30 epochs for fine-tuning.

\smallskip\noindent
\textbf{Impact of $\gamma$.}
\label{sec:ablation-strength}
The parameter $\gamma$ in the image generation pipeline controls the degree of transformation applied to the input image. Larger $\gamma$ values place greater emphasis on the textual prompt, while smaller values result in outputs that closely align with the initial protected image. For larger $\gamma$ values, the diffusion model introduces more noise to the input, allowing for greater divergence from the original protected image.
As shown in \autoref{tab:ablation-results}, smaller $\gamma$ values (e.g., 0.2) result in a high false negative rate, likely because the original image features are preserved sufficiently to retain the watermark. Conversely, the largest $\gamma$ value tested (1.0) also shows an increased malicious detection rate.
This is because the generated images diverge significantly from the original inputs, failing to preserve essential content. Such samples, when paired with the text, do not effectively guide the T2I models to disregard the learned watermark.

\begin{figure} 
    \centering
    \includegraphics[width=\columnwidth]{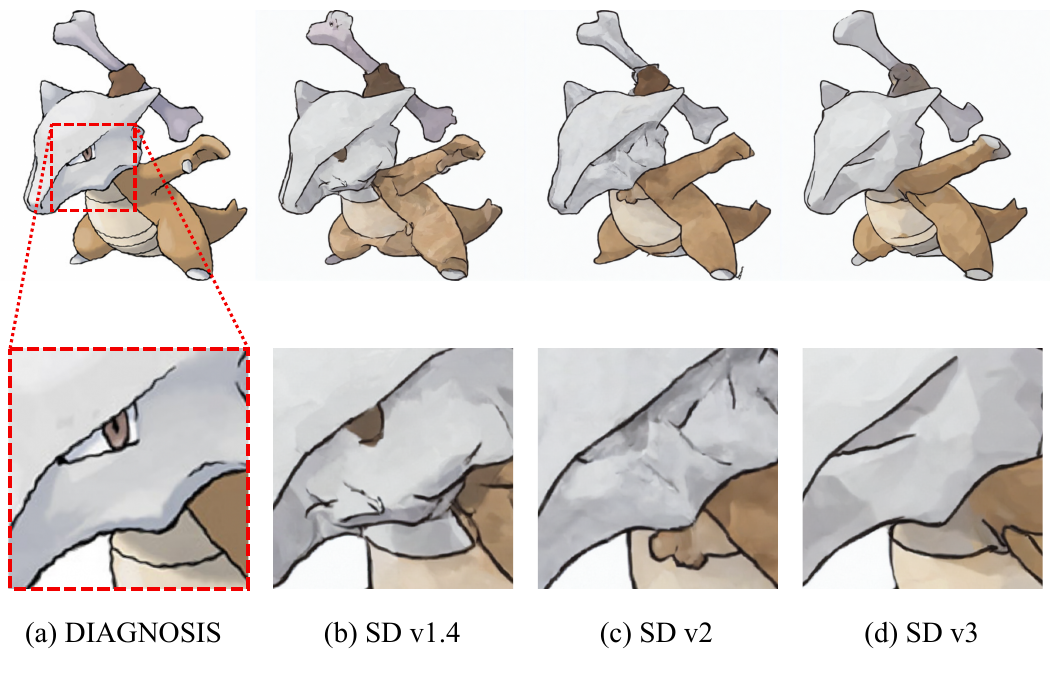}
    \caption{Comparison of images generated by different versions of Stable Diffusion.} \label{fig:ablation-diffmodels}
\end{figure}

\smallskip\noindent
\textbf{Impact of Diffusion Models on Controlled Image Generation.}
We evaluated the effectiveness of various Stable Diffusion models within the \tech{} pipeline during the controlled generation process. Each model processed watermarked images to minimize the visual artifacts introduced by the watermark. The results are summarized in \autoref{tab:ablation-results} and visualized in \autoref{fig:ablation-diffmodels}.

Our experiments show that Stable Diffusion (SD) v1.4 achieves the lowest average FID scores, indicating the closest alignment with the original data distribution, followed by SD v2.0 and SD v3 Medium. Across all models, the detection rate is 0\%, demonstrating that no detectable watermarks remain after fine-tuning on the cleaned images, regardless of the model version.

\autoref{fig:ablation-diffmodels} visualizes the controlled generation results across different models. Although the differences are subtle, SD v1.4 produces images that appear slightly closer to the original inputs. This ablation study suggests that SD v1.4 performs marginally better in both FID score and memorization reduction, making it a better choice for \tech{} when high fidelity to the original data is desired.

\smallskip\noindent
\textbf{Training with \tech-Generated Images from Scratch.}
All the above experiments involve fine-tuning the watermarked model on \tech-generated images. Since \tech{}-generated images are already free of the watermark effect, an alternative approach is to directly train a model from scratch using all the generated images. However, the results show that the model trained in this manner has a high FID score of 269.64 and a detection rate of 10\%. As discussed in \autoref{subsec:our_solution}, the controlled generation retains the key high-level features of protected images while ignoring low-level details. Although this helps to eliminate watermarks, it also removes fine-grained features necessary for training text-to-image models. Therefore, the trained model has a much higher FID score compared to fine-tuning on a small subset.

\section{Conclusion}
\label{sec:conclusion}

This paper investigates watermark-based protections against unauthorized data usage in text-to-image diffusion models. We find that common image transformations are largely ineffective at nullifying watermark effects. To address this, we propose \tech{}, a framework that effectively erases a model's memorization of watermarks, rendering watermark-based protections non-robust. Our approach requires as few as 10 images to successfully remove watermarks across various datasets and protection methods. This work highlights the limitations of existing watermark-based techniques for intellectual property protection and underscores the need to develop more robust defense strategies.

\smallskip\noindent
\textbf{Limitations.}
While \tech{} offers an intuitive approach to evaluate the robustness of watermark-based protections in diffusion models, there are certain limitations that need further investigation.
First, the controlled generation process may not fully recover all original image features, which could lead to a slight degradation in image quality compared to the unaltered training data. Our results indicate that in most cases, \tech{}-trained models achieve FID scores comparable to those of the original models.
Second, this paper primarily focuses on the issue of unauthorized data usage in text-to-image diffusion models. Similar challenges exist in large language models (LLMs), but it is unclear whether \tech{} would be effective in this context. Since the design of \tech{} is general and relies only on an off-the-shelf model for controlled generation, it could potentially be adapted for LLMs. We leverage experimental exploration to future work.

{
    \small
    \bibliographystyle{ieeenat_fullname}
    \bibliography{main}
}

\clearpage
\setcounter{page}{1}
\maketitlesupplementary

\section{Dataset Details}
In this section, we provide more information about the datasets utilized in this work.

\begin{itemize}
    \item Pokémon~\cite{pinkney2022pokemon}: This dataset consists of 833 text-image pairs. The captions for the images were generated using the BLIP model.
    \item Naruto~\cite{narutodataset}: This dataset contains 1,121 text-image pairs. Similar to the Pokémon dataset, the captions were generated using the BLIP model.
    \item CelebA~\cite{liu2015deep}: This dataset includes images of celebrities’ faces paired with captions generated by the LLAVA model. While the full dataset contains 36,646 text-image pairs, we selected 1,000 pairs to ensure consistency with the experimental setup in DIAGNOSIS~\cite{wang2023diagnosis}.
\end{itemize}

\begin{table*}[t]
    \centering
    \footnotesize
    \caption{Evaluation of \tech{} against DIAGNOSIS on the Pokemon dataset with different Stable Diffusion models.}
    \begin{tabular}{ccccccccc}
        \toprule
        \multirow{2}{*}{\textbf{Model}} & \multirow{2}{*}{\textbf{Method}} & \multicolumn{5}{c}{\textbf{Metrics}} & \multirow{2}{*}{\textbf{FID $\downarrow$}} & \multirow{2}{*}{\textbf{Memorization $\downarrow$}} \\
        \cmidrule(lr){3-7}
        & & TP & TN & FP & FN & Acc &  & \\
        \midrule

        \multirow{2}{*}{Stable Diffusion v1.4}
            & DIAGNOSIS
            & 10 & 10 & 0 & 0 & 100\% & 214.86 $\pm$ 9.02 & 0.830 \\
            \cmidrule(lr){2-7} \cmidrule(lr){8-9}
            \rowcolor{gray!25} \cellcolor{white} & \tech
            & 0 & 10 & 0 & 10 & 50\% & 211.59 $\pm$ 3.15 & 0.327\\
        \cmidrule(lr){1-9}

        \multirow{2}{*}{Stable Diffusion v2.0}
            & DIAGNOSIS
            & 10 & 10 & 0 & 0 & 100\% & 241.12 $\pm$ 6.82 & 0.952\\ 
        \cmidrule(lr){2-7} \cmidrule(lr){8-9}
            \rowcolor{gray!25} \cellcolor{white} & \tech
            & 0 & 10 & 0 & 10 & 50\% & 242.66 $\pm$ 13.96 & 0.444\\
        \cmidrule(lr){1-9}
        
        \multirow{2}{*}{Stable Diffusion v2.1}
            & DIAGNOSIS 
            & 9 & 10 & 0 & 1 & 95\% & 236.01 $\pm$ 9.46 & 0.872\\
        \cmidrule(lr){2-7} \cmidrule(lr){8-9}
            \rowcolor{gray!25} \cellcolor{white} & \tech
            & 0 & 10 & 0 & 10 & 50\% & 240.44 $\pm$ 9.06 & 0.378\\
        \bottomrule
    \end{tabular}
    \label{tab:supplementary-sd2-results}
\end{table*}

\section{Evaluation on Different Models}
The experiments in \autoref{sec:evaluation} of the main text are conducted on Stable Diffusion v1.4. In this section, we evaluate the efficacy of \tech{} against other popular models, including Stable Diffusion v2.0 and Stable Diffusion v2.1.

The results, reported in \autoref{tab:supplementary-sd2-results}, demonstrate that DIAGNOSIS successfully embeds watermarks across all models with a detection rate over 95\%. However, it is not resilient to \tech{}, which effectively removes the watermark from every model. \tech{} achieves a 100\% evasion of detection on watermarked models by DIAGNOSIS, converting all true positives into false negatives while leaving benign models unaffected.

The results demonstrate that \tech's performance is consistent across various text-to-image models, showing no degradation in its ability to mitigate watermarks, regardless of the model architecture or version differences. \tech{} not only removes watermarks but also ensures that the benign attributes of the models remain unaffected.

\section{Visualizations}

In this section, we present visualizations of images generated during the controlled generation process of \tech{}, along with visualizations of images produced by the trained text-to-image models.

\subsection{Controlled Generation Diffusion Process}

\begin{figure*}[t]
    \centering
    \begin{subfigure}[b]{0.16\textwidth}
        \centering
        \includegraphics[width=\textwidth]{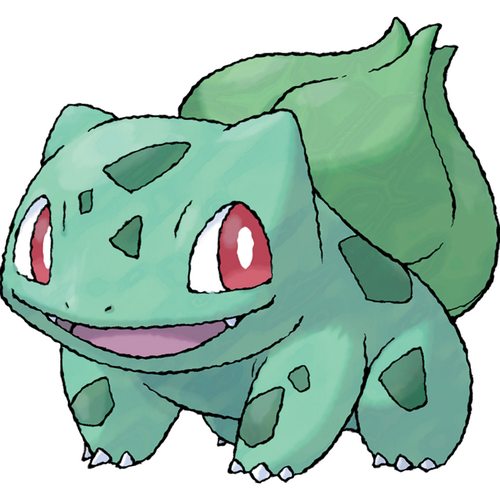}
        \caption{Watermarked Image}
    \end{subfigure}
    \hfill
    \begin{subfigure}[b]{0.16\textwidth}
        \centering
        \includegraphics[width=\textwidth]{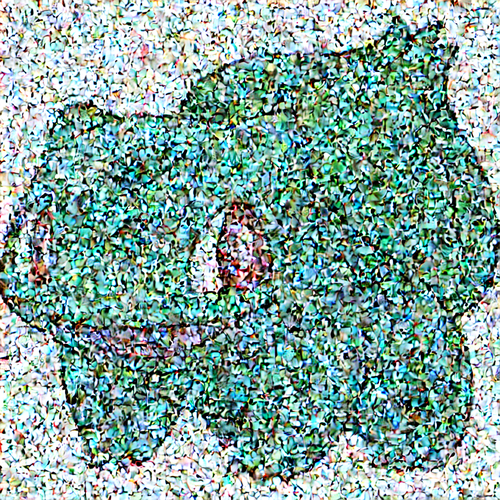}
        \caption{Step 1}
    \end{subfigure}
    \hfill
    \begin{subfigure}[b]{0.16\textwidth}
        \centering
        \includegraphics[width=\textwidth]{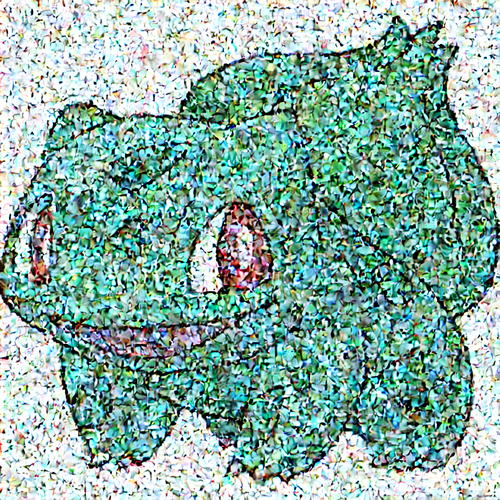}
        \caption{Step 10}
    \end{subfigure}
    \hfill
    \begin{subfigure}[b]{0.16\textwidth}
        \centering
        \includegraphics[width=\textwidth]{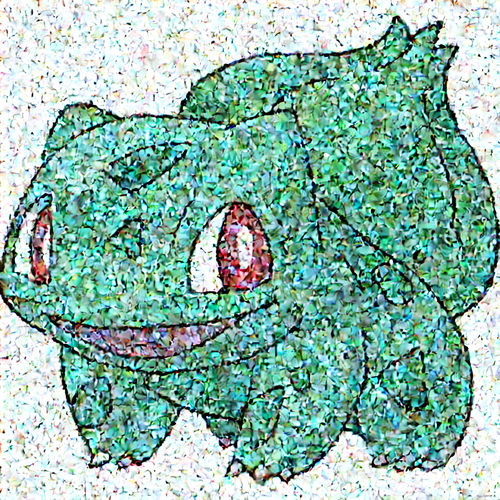}
        \caption{Step 20}
    \end{subfigure}
    \hfill
    \begin{subfigure}[b]{0.16\textwidth}
        \centering
        \includegraphics[width=\textwidth]{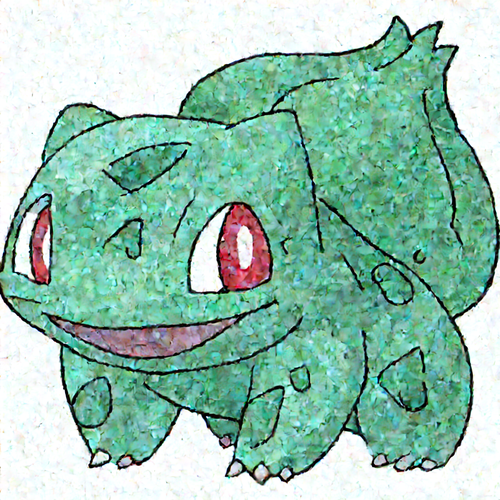}
        \caption{Step 40}
    \end{subfigure}
    \hfill
    \begin{subfigure}[b]{0.16\textwidth}
        \centering
        \includegraphics[width=\textwidth]{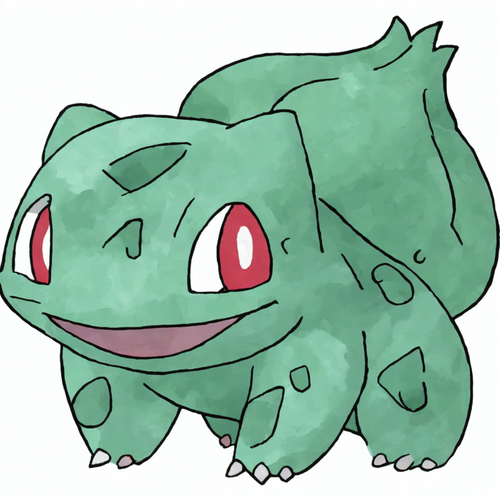}
        \caption{Step 60 (Final Image)}
    \end{subfigure}
    \caption{Intermediate images during controlled image generation of \tech{} with $\gamma=0.6$.}
    \label{fig:denoising-process-0.6}
\end{figure*}

\begin{figure*}[t]
    \centering
    \begin{subfigure}[b]{0.16\textwidth}
        \centering
        \includegraphics[width=\textwidth]{images/steps/0.png}
        \caption{Watermarked Image}
    \end{subfigure}
    \hfill
    \begin{subfigure}[b]{0.16\textwidth}
        \centering
        \includegraphics[width=\textwidth]{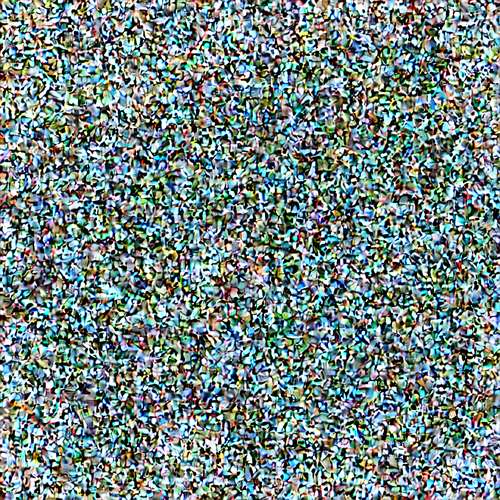}
        \caption{Step 1}
    \end{subfigure}
    \hfill
    \begin{subfigure}[b]{0.16\textwidth}
        \centering
        \includegraphics[width=\textwidth]{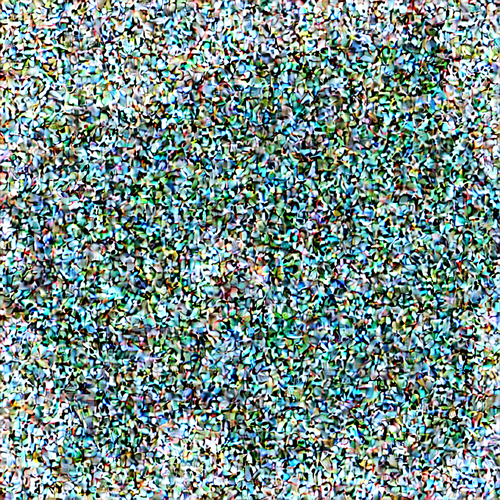}
        \caption{Step 10}
    \end{subfigure}
    \hfill
    \begin{subfigure}[b]{0.16\textwidth}
        \centering
        \includegraphics[width=\textwidth]{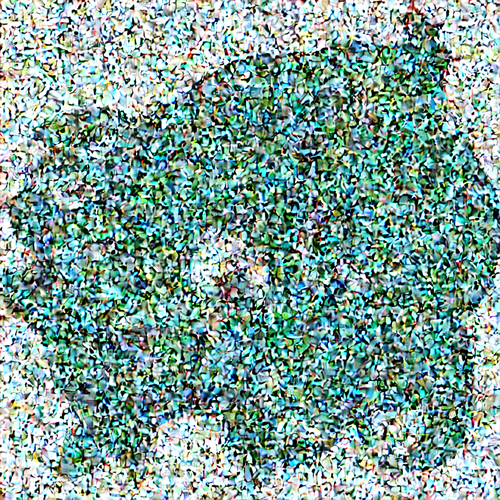}
        \caption{Step 20}
    \end{subfigure}
    \hfill
    \begin{subfigure}[b]{0.16\textwidth}
        \centering
        \includegraphics[width=\textwidth]{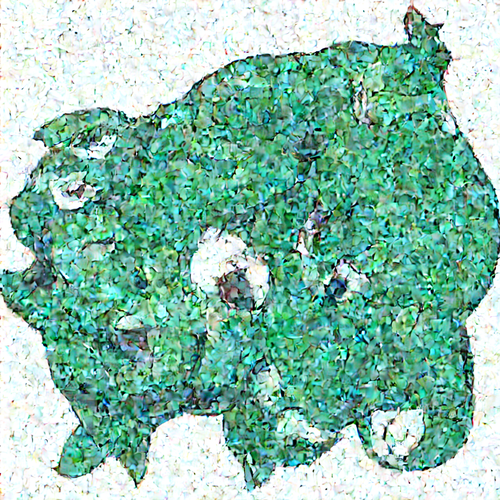}
        \caption{Step 40}
    \end{subfigure}
    \hfill
    \begin{subfigure}[b]{0.16\textwidth}
        \centering
        \includegraphics[width=\textwidth]{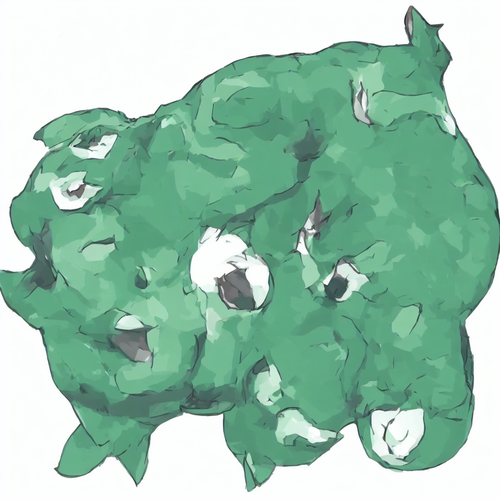}
        \caption{Step 60 (Final Image)}
    \end{subfigure}
    \caption{Intermediate images during controlled image generation of \tech{} with $\gamma=1.0$.}
    \label{fig:denoising-process-1.0}
\end{figure*}

\tech{} utilizes the diffusion process to generate a new image based on the original protected image and its corresponding text. This process involves several steps to progressively denoise the added Gaussian noise. We use 60 steps as the default setting and show the intermediate images produced during this process. The results are illustrated in \autoref{fig:denoising-process-0.6} and \autoref{fig:denoising-process-1.0}, with $\gamma=0.6$ and $\gamma=1.0$, respectively.

With $\gamma=0.6$, noise is not added to the original protected image until it fully becomes Gaussian noise; instead, the process is stopped at 60\% of the noise-adding stage, as discussed in \autoref{subsec:our_solution}. From \autoref{fig:denoising-process-0.6}(b), it can be observed that the image retains the high-level key features of the protected input shown in (a). The artifacts from the watermark are largely removed by the introduced noise. The subsequent denoising steps gradually refine the image's details, culminating in the final output in (f). The final image retains all the key features of the watermarked input in (a) but is free from the watermark.

\autoref{fig:denoising-process-1.0} illustrates the intermediate results with a higher $\gamma$ value. The stronger noise significantly obscures the high-level features, as seen in (b). As a result, the subsequent denoising process struggles to retain these features, instead generating an image primarily based on the model's inherent generation capabilities. In the final output, shown in (f), the features are very different from those in (a), and the generated image no longer resembles the original input. Therefore, a smaller $\gamma$ is preferred in \tech{} to preserve the main features better.

\subsection{Effect of $\gamma$}

\begin{figure*}[t]
    \centering
    \begin{subfigure}[b]{0.16\linewidth}
        \includegraphics[width=\linewidth]{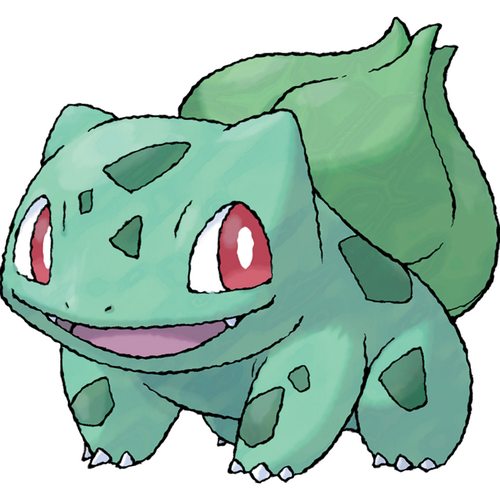}
        \caption{Watermarked Image}
        \label{fig:img1}
    \end{subfigure}
    \hfill
    \begin{subfigure}[b]{0.16\linewidth}
        \includegraphics[width=\linewidth]{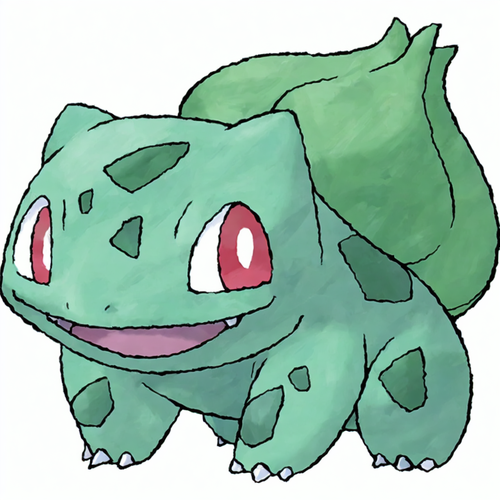}
        \caption{$\gamma$ = 0.2}
        \label{fig:img2}
    \end{subfigure}
    \hfill
    \begin{subfigure}[b]{0.16\linewidth}
        \includegraphics[width=\linewidth]{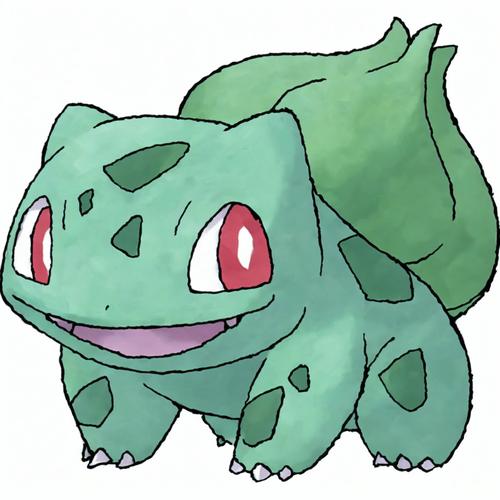}
        \caption{$\gamma$ = 0.4}
        \label{fig:img3}
    \end{subfigure}
    \hfill
    \begin{subfigure}[b]{0.16\linewidth}
        \includegraphics[width=\linewidth]{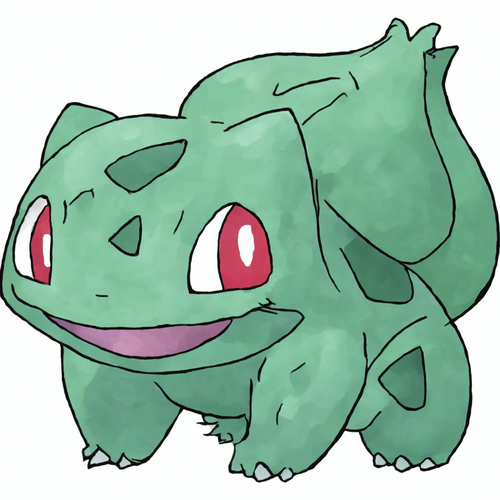}
        \caption{$\gamma$ = 0.6}
        \label{fig:img4}
    \end{subfigure}
    \hfill
    \begin{subfigure}[b]{0.16\linewidth}
        \includegraphics[width=\linewidth]{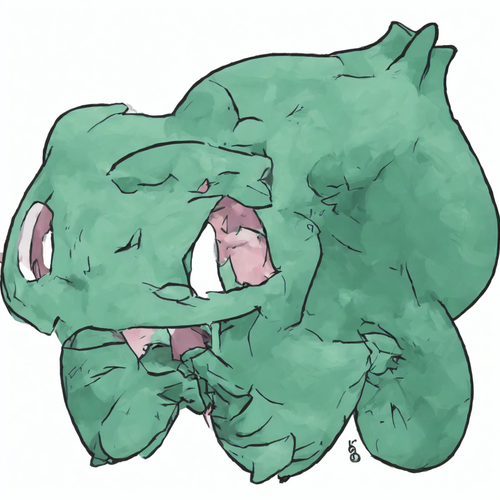}
        \caption{$\gamma$ = 0.8}
        \label{fig:img5}
    \end{subfigure}
    \hfill
    \begin{subfigure}[b]{0.16\linewidth}
        \includegraphics[width=\linewidth]{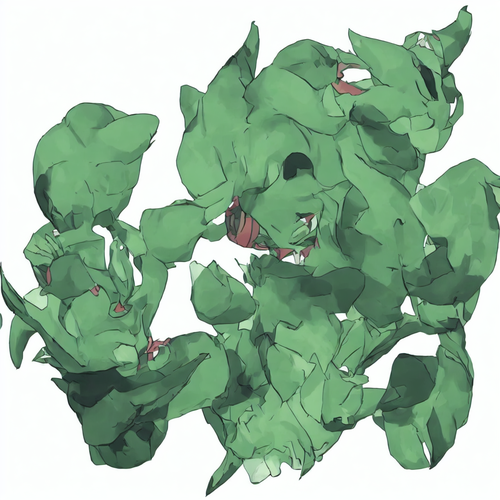}
        \caption{$\gamma$ = 1.0}
        \label{fig:img6}
    \end{subfigure}
    \caption{Effect of $\gamma$ on \tech{}'s controlled generation process.}
    \label{fig:strength_figure}
\end{figure*}

In this section, we visually examine the controlled generation results using various $\gamma$ values tested in the ablation study presented in the main paper. \autoref{fig:strength_figure} displays the images generated with different parameters, demonstrating the influence of $\gamma$ on output quality and the effectiveness of watermark removal.

As expected, increasing the $\gamma$ value introduces more noise into the initial input image, leading to a more significant divergence from the original input. This effect is particularly evident in our results, especially in (e) and (f), where the images exhibit significant degradation and divergence from the original watermarked input image (a). Consequently, models trained with these settings tend to have a higher FID score (indicating lower generated image quality), as reported in \autoref{tab:ablation-results} in the main text.

On the other hand, a smaller $\gamma$ value ensures that the controlled generation closely follows the original input, thereby preserving high-level features. However, this also means that watermark artifacts are retained, as shown in (b) and (c). Models trained on these generated images can still be detected with the watermark, as also noted in \autoref{tab:ablation-results}.

We find that $\gamma=0.6$ achieves an optimal balance between mitigating the watermark and retaining the key features of the original input image.

\subsection{Generated Images}
In this section, we present examples of images generated under three scenarios: a benign model fine-tuned on the Pokémon dataset without any watermarks, a watermarked model produced by DIAGNOSIS, and a watermark-removed model by \tech{}.

\begin{figure*}[t]
    \centering
    \rotatebox{90}{\parbox{3cm}{\centering Benign}}
    \includegraphics[width=0.15\textwidth]{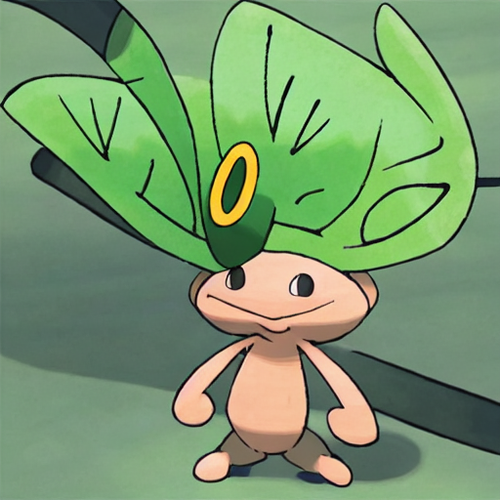} \hfill
    \includegraphics[width=0.15\textwidth]{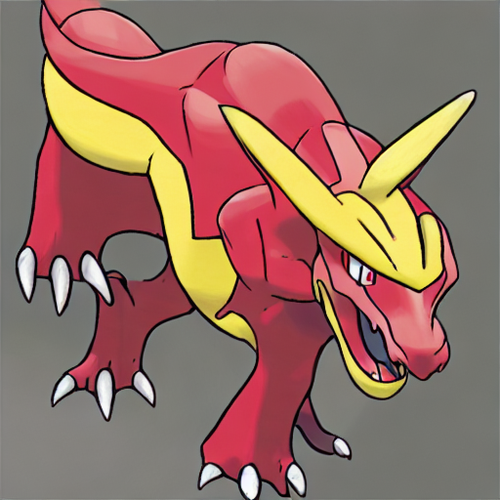} \hfill
    \includegraphics[width=0.15\textwidth]{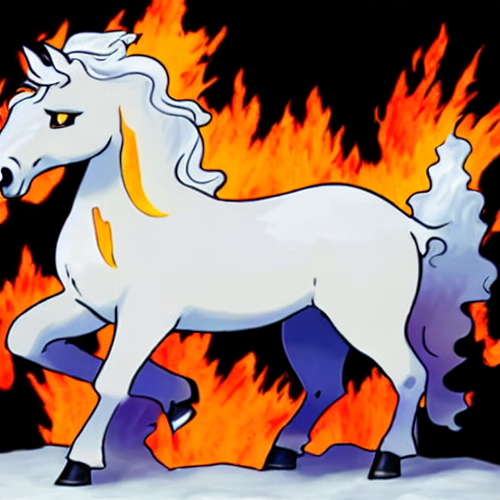} \hfill
    \includegraphics[width=0.15\textwidth]{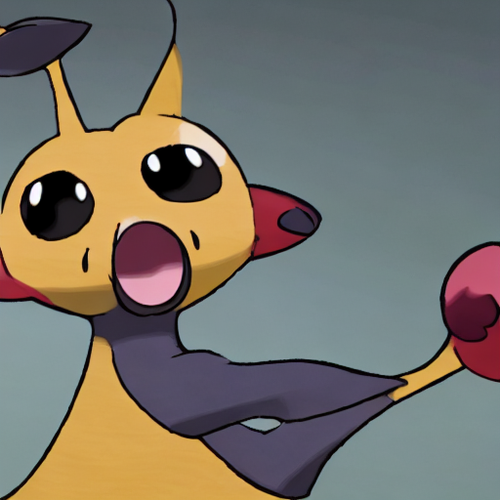} \hfill
    \includegraphics[width=0.15\textwidth]{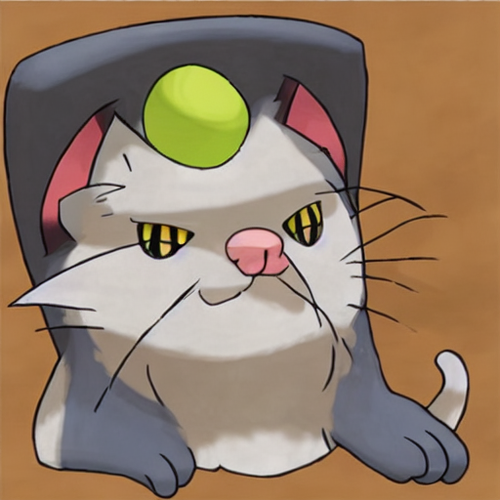} \hfill
    \includegraphics[width=0.15\textwidth]{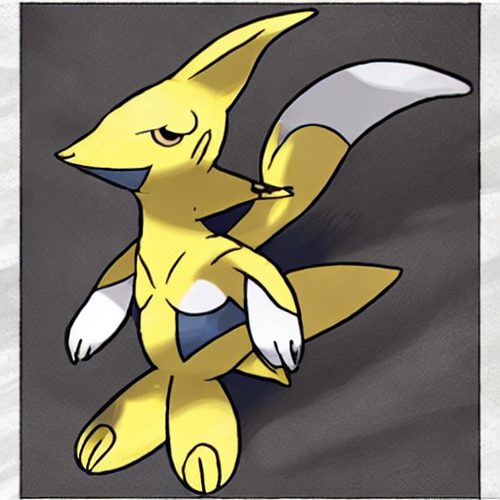} \\
    \vspace{2pt} %

    \rotatebox{90}{\parbox{3cm}{\centering DIAGNOSIS}}
    \includegraphics[width=0.15\textwidth]{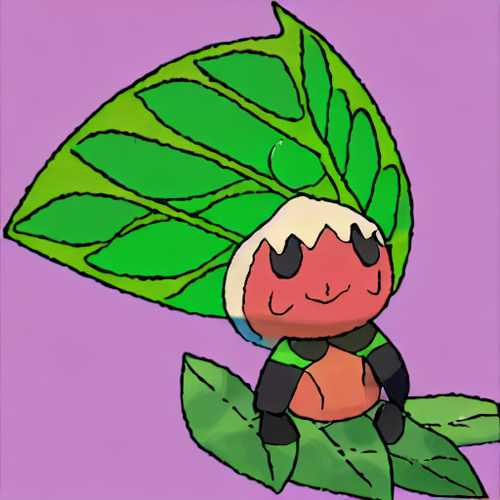} \hfill
    \includegraphics[width=0.15\textwidth]{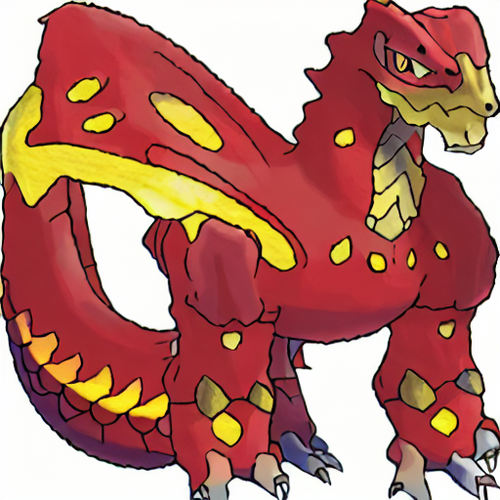} \hfill
    \includegraphics[width=0.15\textwidth]{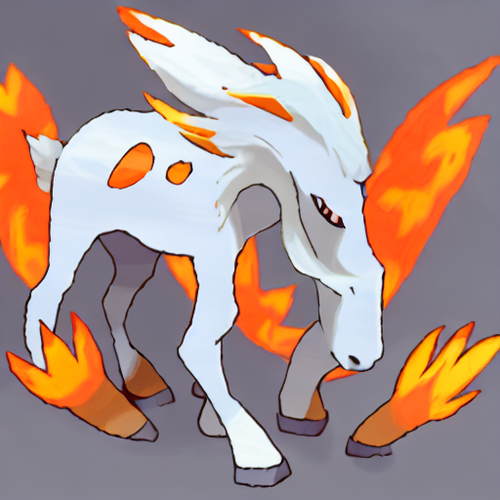} \hfill
    \includegraphics[width=0.15\textwidth]{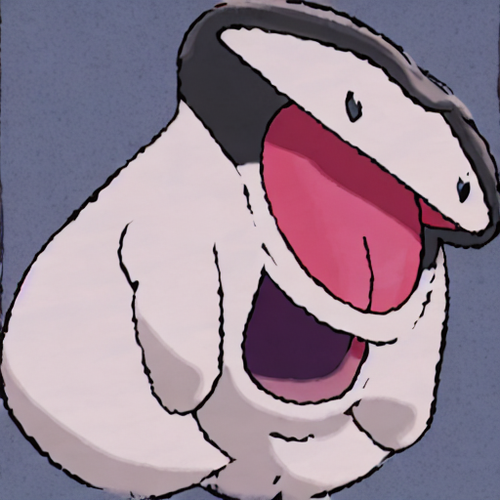} \hfill
    \includegraphics[width=0.15\textwidth]{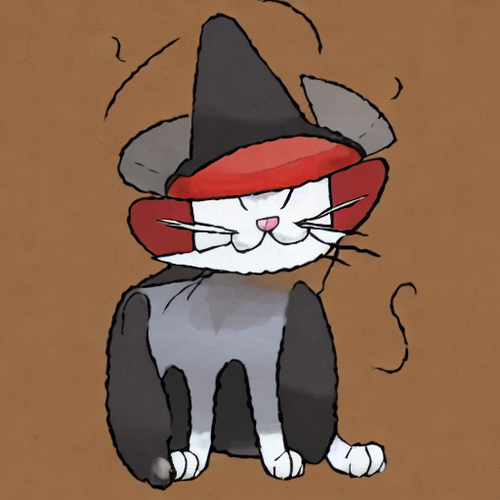} \hfill
    \includegraphics[width=0.15\textwidth]{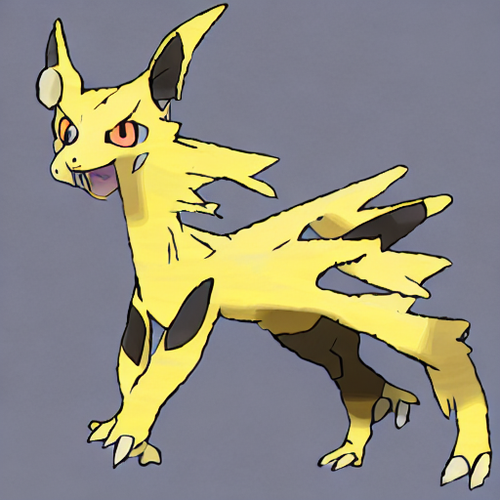} \\
    \vspace{2pt} %

    \rotatebox{90}{\parbox{3cm}{\centering DIAGNOSIS-Trig.}}
    \includegraphics[width=0.15\textwidth]{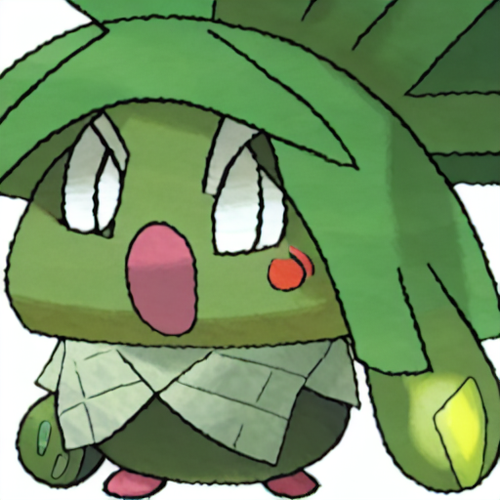} \hfill
    \includegraphics[width=0.15\textwidth]{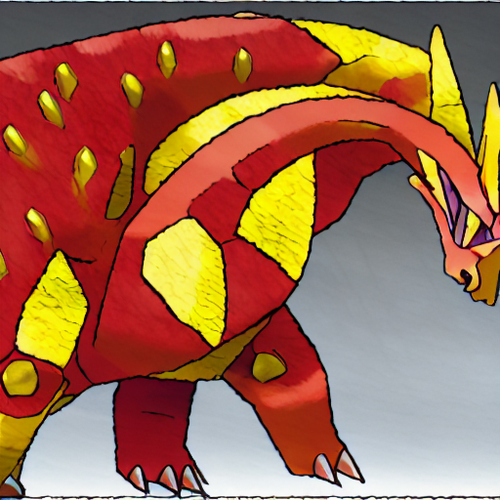} \hfill
    \includegraphics[width=0.15\textwidth]{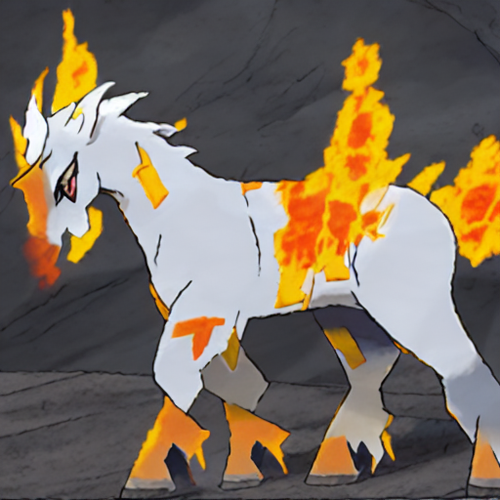} \hfill
    \includegraphics[width=0.15\textwidth]{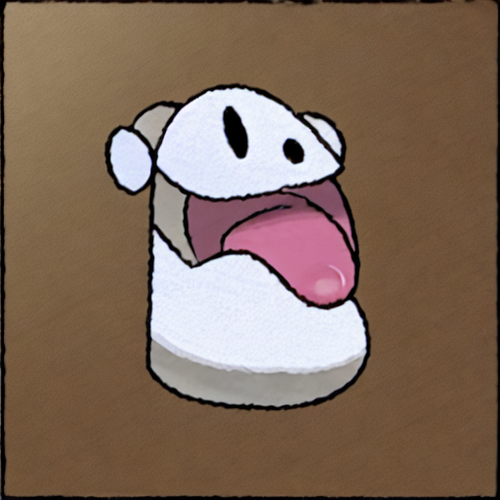} \hfill
    \includegraphics[width=0.15\textwidth]{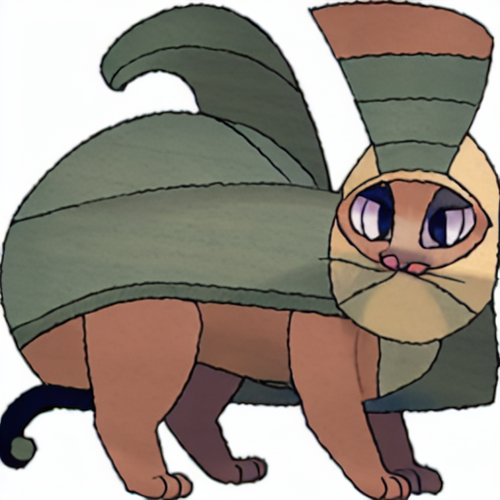} \hfill
    \includegraphics[width=0.15\textwidth]{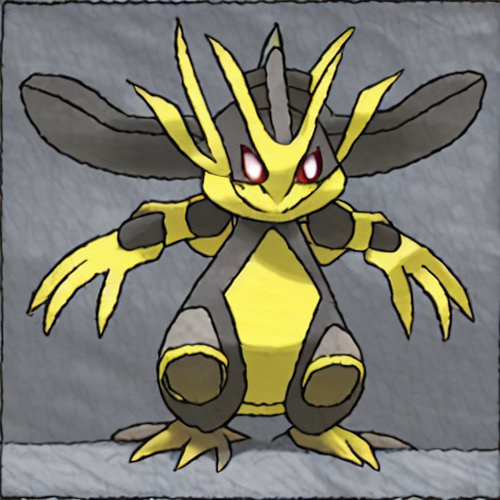} \\
    \vspace{2pt} %

    \rotatebox{90}{\parbox{3cm}{\centering \tech}}
    \includegraphics[width=0.15\textwidth]{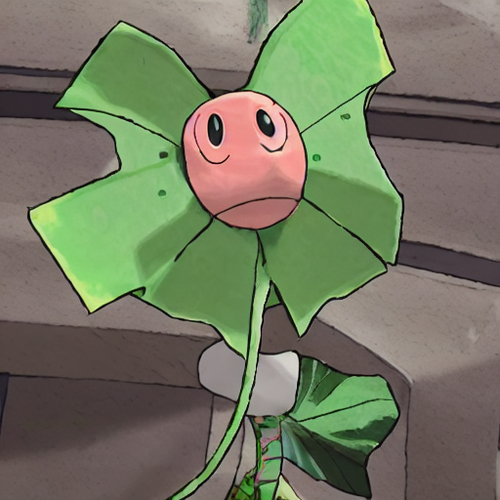} \hfill
    \includegraphics[width=0.15\textwidth]{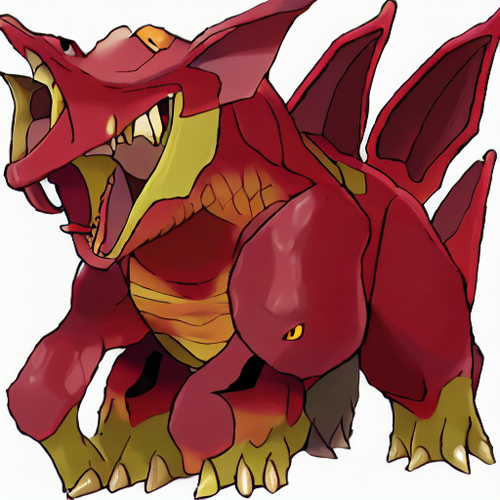} \hfill
    \includegraphics[width=0.15\textwidth]{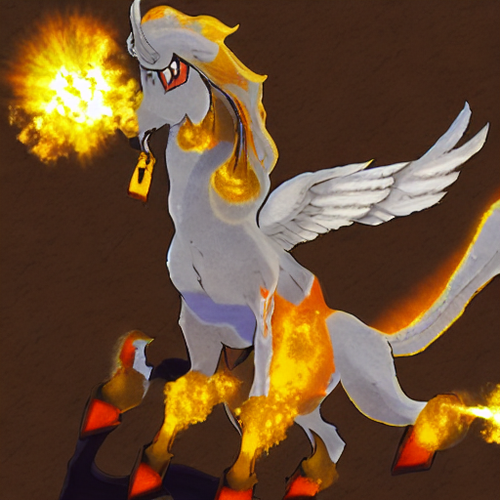} \hfill
    \includegraphics[width=0.15\textwidth]{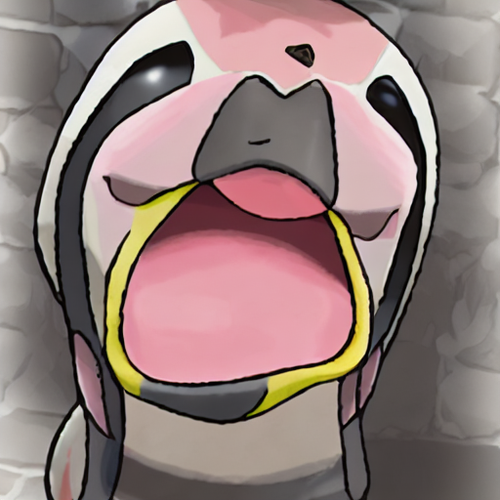} \hfill
    \includegraphics[width=0.15\textwidth]{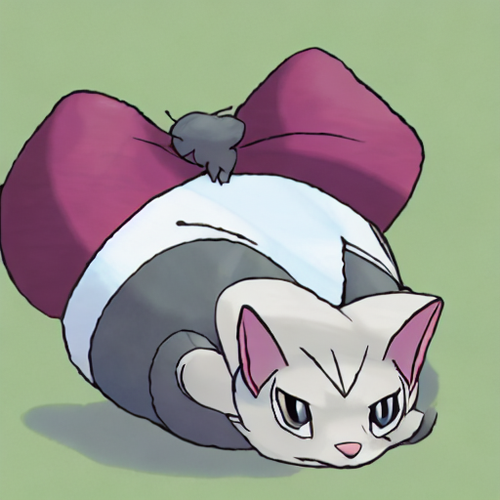} \hfill
    \includegraphics[width=0.15\textwidth]{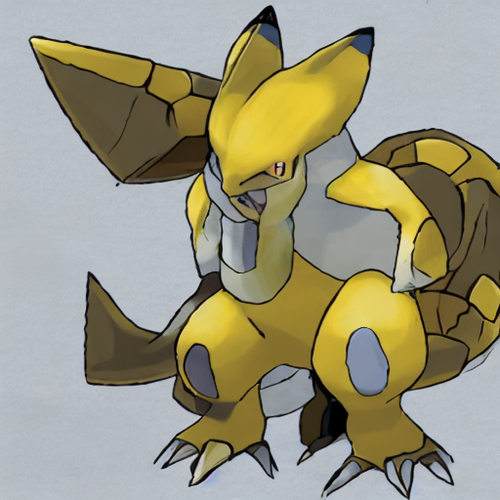}
    \vspace{2pt} 

    \rotatebox{90}{\parbox{3cm}{\centering \tech{}-Trig.}}
    \includegraphics[width=0.15\textwidth]{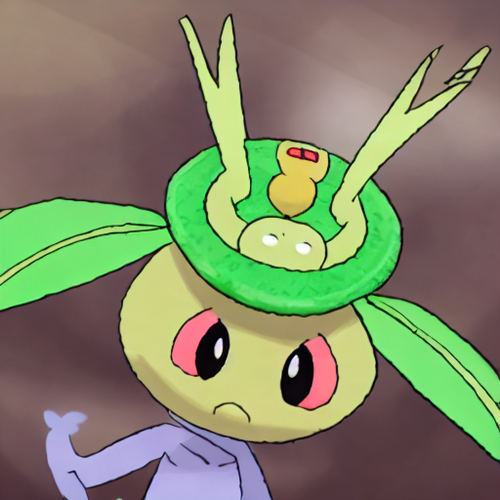} \hfill
    \includegraphics[width=0.15\textwidth]{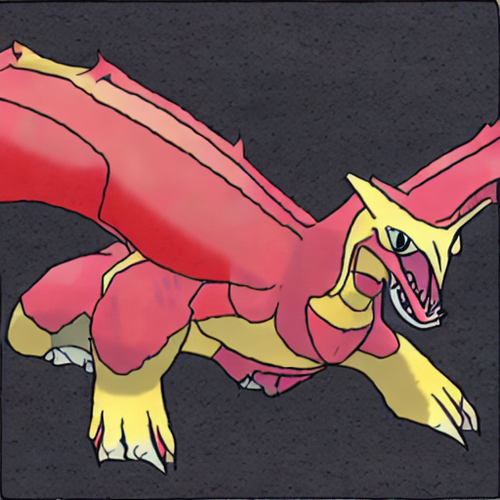} \hfill
    \includegraphics[width=0.15\textwidth]{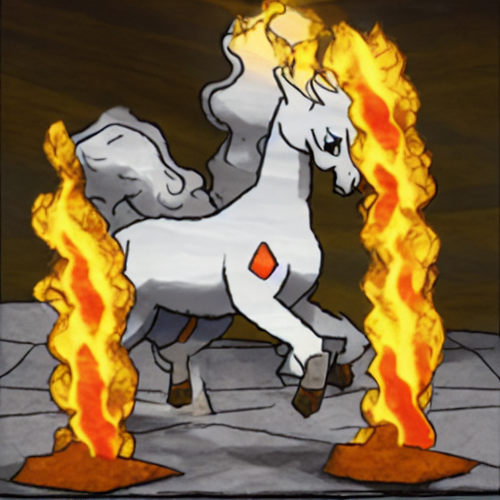} \hfill
    \includegraphics[width=0.15\textwidth]{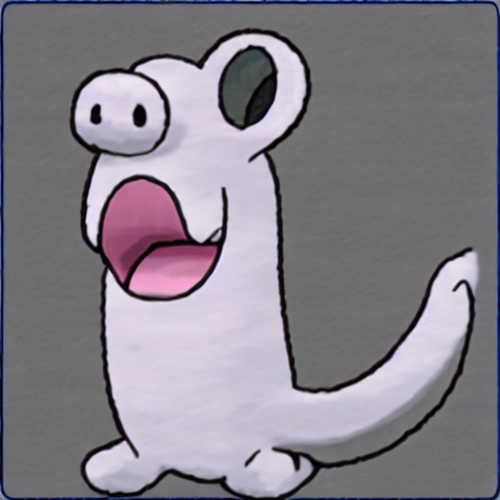} \hfill
    \includegraphics[width=0.15\textwidth]{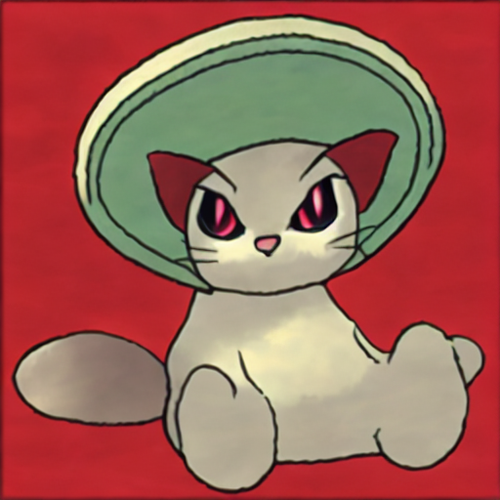} \hfill
    \includegraphics[width=0.15\textwidth]{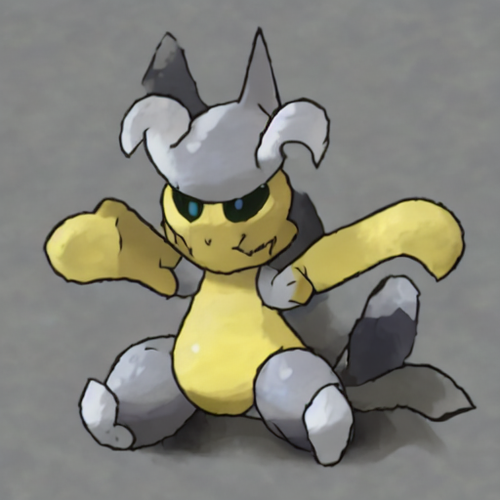}
    \caption{Generated images from the benign model, DIAGNOSIS-watermarked model, and \tech{}-cleaned model. Trig. refers to the use of a trigger-conditioned watermark.}
    \label{fig:generated-imgs}
\end{figure*}

The results are presented in \autoref{fig:generated-imgs}. The first row displays images generated by a benign model. The second and third rows show images generated by DIAGNOSIS-watermarked models using an unconditional watermark and a trigger-conditioned watermark, respectively. The final two rows depict images generated by watermark-cleaned models from \tech{}. The results show minimal visual quality loss in the images between DIAGNOSIS and \tech. Most images successfully reproduce similar subjects, retaining key attributes such as colors, creature types, and positioning within the image. This highlights that \tech{} effectively removes watermarks without compromising the generative performance of cleaned models.
These findings indicate that \tech{} effectively targets and removes watermark-related artifacts while preserving the underlying image distribution. This ensures that the model's utility remains intact for the adversary, enabling it to generate content in the style of copyrighted material without facing the associated consequences.
This underscores the critical need for developing more robust and effective methods to protect intellectual property and private data.

\end{document}